\documentclass{article}
\usepackage{spconf,amsmath,graphicx,amsfonts,xcolor,comment,caption, cleveref,multirow, multicol}
\usepackage{tikz}
\usetikzlibrary{shapes,arrows,positioning}

\title{Balancing Stability and Plasticity in Sequentially Trained \\ Early-Exiting Neural Networks}
\name{Alaa Zniber$^{{\star}}$, Ouassim Karrakchou$^{{\star}}$, Mounir Ghogho$^{{\dagger}{\ddagger}}$ \thanks{This research has received funding from the European Union's Horizon
research and innovation program under grant agreement No 101070374.}}
\address{$^{\star}$ \textit{TICLab, International University of Rabat, Morocco}\\
$^{\dagger}$ \textit{College of Computing, University Mohammed VI Polytechnic, Morocco}\\
$^{\ddagger}$ \textit{School of Electronic and Electrical Engineering, University of Leeds, UK}}
\begin{document}
%
\maketitle
\begin{abstract}
Early-exiting neural networks enable adaptive inference by allowing inputs to exit at intermediate classifiers, reducing computation for easy samples while maintaining high accuracy. In practice, exits can be trained sequentially by incrementally adding them to a shared backbone; however, this sequential training can cause newly introduced exits to interfere with previously learned ones, degrading the performance of earlier classifiers. We address this problem by retaining the knowledge embedded in existing exits while allowing new ones to specialize. We propose two alternative approaches that operate at different levels of the model. The first constrains learning by protecting parameters that are important for previously trained exits, while the second preserves the output distributions of earlier exits as the network adapts. These alternatives directly reflect the stability-plasticity trade-off studied in continual learning. Accordingly, we leverage \textit{Elastic Weight Consolidation} to constrain critical weights and \textit{Learning without Forgetting} to preserve output distributions. Experiments on standard benchmarks show that our approaches consistently improve early-exit performance, achieving higher accuracy over existing sequential training methods and significant performance speedups at low computational budgets.
\end{abstract}
\section{Introduction}
\label{sec:intro}

The increasing deployment of deep learning models under strict latency, energy, and memory constraints has motivated the development of resource-efficient dynamic architectures. Rather than executing a fixed computation graph for all inputs, such architectures adapt their computational cost to input difficulty, enabling efficient inference while maintaining high accuracy \cite{dynn_survey}. Early-exiting neural networks (EENNs) are a representative instance of this paradigm. EENNs augment a backbone network with multiple internal classifiers (ICs) placed at increasing depths, allowing inference to terminate early once sufficient confidence is reached. Easy samples can be classified using shallow representations, while more challenging inputs propagate deeper into the network. This exit-based adaptivity significantly reduces average inference cost and makes EENNs well-suited for resource-constrained applications \cite{resource_constraint_eenn}.

To realize these benefits in practice, a critical design choice concerns how to train the multiple exits within an EENN. Most prior work has focused on joint training strategies, which optimize the backbone along with all exits simultaneously in an end-to-end manner. Joint training has shown to deliver strong empirical performance and has therefore become the predominant approach \cite{branchynet,bdl,eccv24,icml25}. In contrast, sequential training performs the training incrementally as exits are added to an existing backbone. Sequential training naturally aligns with pretrained backbones by enabling a more rapid design time and an efficient analysis of where meaningful predictions emerge within the network -- a useful property for large models where the computational cost of full joint training can be prohibitive or even undesirable \cite{gaml,berxit}. 

Sequential training approaches for EENNs are categorized into three main groups based on the specific training methodology employed for each successive exit \cite{icml25}. \textbf{Disjoint training} (also referred to as two-stage training) first trains the backbone independently and then freezes it while learning ICs \cite{ztw,lgvit}. This approach is particularly effective for retrofitting pretrained models with early exits at the cost of limited adaptability; new exits cannot refine earlier representations to better support their specific objectives. \textbf{Branch-wise training} introduces exits sequentially while unfreezing only the backbone segment associated with the current exit, offering moderate flexibility and helping to alleviate vanishing or exploding gradients in deep architectures \cite{msdnet}. However, this forward-thinking approach suffers from premature freezing: once an exit is trained and the training process moves deeper, that exit's parameters become fixed, preventing further refinement even as later exits reveal useful representational structures. \textbf{Separate training} further relaxes these constraints by unfreezing the full network \cite{icml25}. The exits are also introduced sequentially, however the training of each exit is done jointly with previous exits, allowing each branch to capture features at different abstraction levels through potentially distinct objectives. Although this approach provides greater flexibility, it applies uniform updates to all backbone parameters during each iteration, without identifying which parameters are essential for earlier exits and which can be modified for new ones.

A key challenge unifies these methods: introducing new exits disrupts previously learned representations through gradient interference, with degradation intensifying as more exits are added \cite{survey_eenn}. This results from current approaches disregarding the heterogeneous importance of parameters, particularly those that have already been tuned for earlier exits. Indeed, some weights encode critical features that earlier exits depend on, while others are less consequential and could be freely adapted. Therefore, distinguishing between these parameter types is essential as it enables selective protection of critical knowledge while maintaining the plasticity needed for new exits to specialize effectively. Importantly, this challenge is not unique to early-exit networks. In continual learning (CL), models face a similar dilemma when learning sequential tasks; they must acquire new capabilities (plasticity) while preserving previously learned knowledge (stability) to mitigate catastrophic forgetting \cite{survey_cl}. We thus recognize that sequential early-exit training exhibits this same stability-plasticity trade-off. This conceptual alignment suggests that CL techniques can be adapted to protect salient features of earlier exits from being forgotten during sequential training. It is worth noting that the goal here is not to implement CL on EENN (e.g., Szatkowski et al. \cite{eenn_cl}), but to benefit from some of the regularization techniques of CL to mitigate representation degradation in sequential training of EENN.

Building on this perspective, we propose a novel two-stage sequential training approach. In the first stage, we pretrain the backbone using only the final exit in order for deep layers to develop rich abstract representations before early exits are introduced \cite{deebert,fastbert}. In the second stage, we perform sequential training by explicitly preserving the knowledge acquired by earlier exits as new exits are introduced. Consistent with the CL perspective, our goal is to counteract the forgetting induced by gradient interference. Without protection mechanisms, optimizing a new exit inevitably overwrites representations that previous exits depend on, leading to progressive degradation. Therefore, we aim to regularize sequential training from two alternative standpoints, namely at the parameter and output distribution levels, each addressing forgetting from a different aspect. At the parameter level, forgetting occurs because optimization updates treat all weights equally, although some are critical for the performance of existing exits. By identifying which parameters are essential, we can selectively protect them, allowing the network to adapt safely while preserving knowledge that earlier exits rely on. This motivates the use of Elastic Weight Consolidation (EWC) \cite{ewc_first,ewc_online}, which estimates parameter importance via Fisher information and constrains updates on crucial weights, thus weighting the learning process to balance stability and plasticity. At the output distribution level, however, what ultimately matters is that the predictive behavior of the earlier exits remains stable as new exits are trained. To achieve this, we adopt Learning without Forgetting (LwF) \cite{LwF}, which regularizes the outputs of earlier exits to match their original predictions, maintaining functional consistency as new exits are trained. Both EWC and LwF are governed by tunable hyperparameters, giving explicit control over the trade-off between preserving early-exit performance and enabling adaptation for new exits.

The main contributions of this work are threefold:

\begin{itemize}
    \item We propose a novel technique to mitigate the degradation of previously learnt representation in order to address the stability-plasticity trade-off faced during sequential training of EENN.
    \item We adapt two CL regularization strategies, EWC for parameter-level protection and LwF for output distribution-level consistency, to enable principled sequential training that balances exit specialization with knowledge preservation.
    \item We demonstrate consistent improvements over existing sequential training methods across benchmark architectures, validating the effectiveness of the CL approach for EENNs.
\end{itemize}

\section{Proposed Methodology}
\label{sec:method}

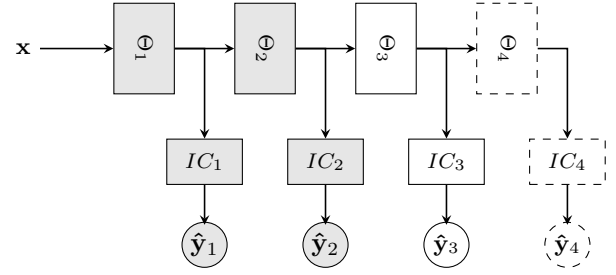
\begin{figure}[t]
\begin{minipage}[b]{\linewidth}
  \centering
  \centerline{\begin{tikzpicture}[node distance=1.6cm]
\tikzstyle{inout} = [text centered, font=\small]
\tikzstyle{nn} = [rectangle, minimum width=1.2cm, minimum height=0.8cm, text centered, draw=black, font=\footnotesize]
\tikzstyle{ic} = [rectangle, minimum width=1cm, minimum height=0.6cm, text centered, draw=black, font=\footnotesize]
\tikzstyle{line} = [semithick,->,>=stealth]
\node(in) [inout] {$\mathbf{x}$};
\node(block1) [nn, right of=in, rotate=-90, fill=gray!20] {$\Theta_1$};
\node(block2) [nn, right of=block1, rotate=-90, fill=gray!20] {$\Theta_2$};
\node(block3) [nn, right of=block2, rotate=-90] {$\Theta_3$};
\node(block4) [nn, right of=block3, rotate=-90, dashed] {$\Theta_4$};

\node(IC1) [ic, below of=block1, xshift=0.8cm, yshift=0.1cm, fill=gray!20] {$IC_1$};
\node(IC2) [ic, below of=block2, xshift=0.8cm, yshift=0.1cm, fill=gray!20] {$IC_2$};
\node(IC3) [ic, below of=block3, xshift=0.8cm, yshift=0.1cm] {$IC_3$};
\node(IC4) [ic, below of=block4, xshift=0.8cm, yshift=0.1cm, dashed] {$IC_4$};

\node(M0hat) [inout, below of=IC1, yshift=0.5cm] {$\mathbf{\hat{y}}_1$};
\node(M1hat) [inout, below of=IC2, yshift=0.5cm] {$\mathbf{\hat{y}}_2$};
\node(M2hat) [inout, below of=IC3, yshift=0.5cm] {$\mathbf{\hat{y}}_3$};
\node(M3hat) [inout, below of=IC4, yshift=0.5cm] {$\mathbf{\hat{y}}_4$};

\node[draw,circle,inner sep=6pt, fill=gray!20] at (M0hat.center) {};
    \node at (M0hat.center) {$\mathbf{\hat{y}}_1$}; 
\node[draw,circle,inner sep=6pt, fill=gray!20] at (M1hat.center) {};
    \node at (M1hat.center) {$\mathbf{\hat{y}}_2$}; 
\node[draw,circle,inner sep=6pt] at (M2hat) {};
\node[draw,circle,inner sep=6pt,dashed] at (M3hat) {};

\draw [line] (in) -- (block1);
\draw [line] (block1) -- (block2);
\draw [line] (block2) -- (block3);
\draw [line] (block3) -- (block4);

\draw [line] (block1.north) -| (IC1.north);
\draw [line] (block2.north) -| (IC2.north);
\draw [line] (block3.north) -| (IC3.north);
\draw [line] (block4.north) -| (IC4.north);

\draw [line] (IC1.south) -- (M0hat.north);
\draw [line] (IC2.south) -- (M1hat.north);
\draw [line] (IC3.south) -- (M2hat.north);
\draw [line] (IC4.south) -- (M3hat.north);

\end{tikzpicture}}
\end{minipage}
  \caption{Flow of the proposed sequential training scheme. Grey refers to regularized parameters or output distributions; dashed lines refer to frozen segments. The backbone is represented by rotated blocks $\Theta_{\mu}$'s and intermediate classifiers by horizontal blocks $IC_{\mu}$'s. Each IC $\mu$ outputs a distribution $\mathbf{\hat{y}}_{\mu}$.}
  \label{fig:earlyexiting}
\end{figure}

We consider an EENN consisting of $M$ ICs, where each backbone subnetwork $\mu$ is parameterized by $\{\Theta_1, \dots, \Theta_{\mu}\}$ and produces class probability vector $\mathbf{\hat{y}}_{\mu}$ (see \Cref{fig:earlyexiting}), with $\hat{y}_{\mu}(c)$ (for $c \in \{1,\dots,C\}$) being the probability associated with class $c$. We frame the sequential training of ICs as a sequence of learning tasks indexed by $\mu \in \{1, 2, \ldots, M\}$, where task $\mu$ corresponds to training IC $\mu$ while preserving the predictive capabilities of all previously trained ICs $\{1, \ldots, {\mu-1}\}$. The shared backbone parameters must accommodate the objectives of all trained ICs, balancing plasticity (i.e., the ability to refine representations for the current task) against stability (i.e., maintaining performance on earlier tasks).

To address this stability-plasticity trade-off, we augment the standard classification loss with a regularization term that protects previously acquired knowledge. The loss function for training exit $\mu$ takes the general form:
\begin{equation}
\mathcal{L}_{\mathrm{total}}^{(\mu)} =
\mathcal{L}_{\mathrm{CE}}^{(\mu)}
+ s \, \lambda \mathcal{R}_{\mathrm{EWC}}^{(\mu)}
+ (1-s)\, \rho \mathcal{R}_{\mathrm{LwF}}^{(\mu)}
\end{equation}

\noindent where $\mathcal{L}_{\mathrm{CE}}^{(\mu)}$ denotes the cross-entropy loss for exit $\mu$ which depends on $\{\Theta_1, \dots, \Theta_{\mu}\}$ and the parameters of IC $\mu$; $\mathcal{R}_{\mathrm{EWC}}^{(\mu)}$ and $\mathcal{R}_{\mathrm{LwF}}^{(\mu)}$ are regularization terms that mitigate the degradation of earlier exits, corresponding to EWC (parameter-level) and LwF (output-distribution-level), respectively. Their relative strengths are controlled by the hyperparameters $\lambda$ and $\rho$. The binary variable $s \in \{0,1\}$ indicates which regularization strategy is active.

To illustrate the training procedure, \Cref{fig:earlyexiting} shows the status of each subnetwork when training exit 3. At this stage, the first two blocks (parameterized by $\Theta_1$ and $\Theta_2$) and their respective classifiers (producing predictions $\mathbf{\hat{y}}_1$ and $\mathbf{\hat{y}}_2$) have already been trained. We now add IC 3 and therefore, we need to train the full network up to exit 3. We minimize the cross-entropy loss $\mathcal{L}^{(3)}_{\text{CE}}$ augmented with regularization $\mathcal{R}^{(3)}$ applied to the previously trained blocks, either constraining their parameters (EWC) or preserving their predictions (LwF). The final block ($\Theta_4$) and its corresponding classifier remain frozen throughout this stage.

\subsection{Parameter-level Regularization} 

The first approach employs parameter-level regularization through EWC, which protects parameters that are critical for previously trained exits by penalizing changes to important weights. This importance score is computed using Fisher Information \cite{fisher_hessian}. Parameters with large Fisher values correspond to directions in parameter space where small perturbations would strongly increase the cross-entropy loss $\mathcal{L}_{\mathrm{CE}}^{(\nu)}$ on previous exits and are therefore prioritized for protection. 

Following the online EWC framework \cite{ewc_online}, the parameter-level regularization is defined as
\begin{equation}
\mathcal{R}_{\mathrm{EWC}}^{(\mu)} =
\sum_{\theta_k \in \cup_{i=1}^{\mu-1} \Theta_i}
\left(
\sum_{\nu=1}^{\mu-1} F_k^{(\nu)}
\right)
\left(\theta_k - \theta_k^{*}\right)^2
\end{equation}
Here, $\theta_k^{*} \in \mathbb{R}$ denotes the value of parameter $\theta_k \in \cup_{i=1}^{\mu-1} \Theta_i$ after training up to exit $\mu-1$.
The term $F_k^{(\nu)} \in \mathbb{R}$ represents the Fisher information measuring the importance of parameter $\theta_k$ for exit $\nu$. In practice, we use the empirical Fisher information \cite{fisher_compute}, which is computed as:
\begin{equation}
F_k^{(\nu)} =
\frac{1}{N} \sum_{\mathbf{x} \in \mathcal{D}}
\left(
\left.
\frac{\partial}{\partial \theta_k}
\log \hat{y}_{\nu}(\mathcal{C}(\mathbf{x}))
\right|_{\theta_k=\theta_k^*}
\right)^2
\end{equation}
where $\mathcal{D}$ denotes the training dataset, 
$N$ is the total number of training samples, and $\hat{y}_{\nu}(\mathcal{C}(\mathbf{x}))$ is the probability associated with the true label $\mathcal{C}(\mathbf{x})\in \{1, \dots, C\}$ for input $\mathbf{x}$ at exit $\nu$.

\subsection{Output distribution-level Regularization}  
The second proposed approach takes a fundamentally different perspective by operating at the output distribution level. Rather than protecting individual parameters, LwF preserves knowledge by enforcing that the current model maintains the input-output mappings learned by previously trained exits. In contrast to EWC, LwF imposes a functional constraint that is agnostic to specific parameter values and instead focuses on preserving decision boundaries. Under LwF, parameters are free to change substantially as long as the output distributions of previous exits is minimally shifted by the training of this new exit, allowing the network to adapt its internal representations while maintaining stable behavior. Importantly, the LwF loss enables gradient flow through both the shared backbone and the classifier heads of earlier exits; this added flexibility allows earlier classifiers to compensate for moderate backbone drift, potentially mitigating performance degradation. The output distribution-level regularization is thus defined as:
\begin{equation}
\mathcal{R}_{\mathrm{LwF}}^{(\mu)} =
\sum_{\nu=1}^{\mu-1}
\mathbb{KL}
\left(
\mathbf{\hat{y}}_{\nu} \;\|\; \mathbf{\hat{y}}^*_{\nu}
\right)
\end{equation}
where $\mathbf{\hat{y}}_{\nu}$ denotes the output distribution of IC $\nu$ when training exit $\mu$, $\mathbf{\hat{y}}^*_{\nu}$ denotes the output distribution obtained with the training of IC $\nu$, and $\mathbb{KL}(\cdot \| \cdot)$ is the Kullback-Leibler divergence.

\section{Experiments}
\label{sec:exprms}

We evaluate our approach on standard early-exit benchmarks following the experimental setup proposed in \cite{icml25}. Our evaluation focuses on comparing the proposed EWC and LwF regularization strategies against existing sequential training methods across different architectures and computational budgets.

\begin{table}[t!]
\centering
\caption{Top-1 accuracy for ResNet-34 and MSDNet on CIFAR-100 compared to baselines.}
\small
\begin{tabular}{c|c|c|c|c|c}
\hline
Model & Method & $25\%$ & $50\%$ & $75\%$ & $100\%$ \\
\hline

\multirow{6}{*}{ResNet34}
 & Disjoint     & $52.57$ & $67.56$ & $73.49$ & $73.79$ \\
 & Branch-wise  & $64.19$ & $67.03$ & $66.88$ & $66.75$ \\
 & Separate     & $66.25$ & $72.57$ & $72.87$ & $72.82$ \\
 & Joint        & $62.84$ & $72.80$ & $\textbf{74.32}$ & $\textbf{74.17}$ \\
\cline{2-6}
 & EWC           & $70.50$ & $\textbf{74.07}$ & $73.90$ & $73.84$ \\
\cline{2-6}
 & LwF       & $\textbf{70.71}$ & $73.84$ & $73.61$ & $73.55$ \\

\hline\hline

\multirow{6}{*}{MSDNet}
 & Disjoint     & $56.74$ & $63.96$ & $68.59$ & $70.36$ \\
 & Branch-wise  & $58.24$ & $59.73$ & $60.16$ & $59.97$ \\
 & Separate     & $64.90$ & $67.84$ & $69.02$ & $69.21$ \\
 & Joint        & $65.93$ & $72.02$ & $\textbf{74.73}$ & $\textbf{75.86}$ \\
\cline{2-6}
 & EWC       & $65.10$ & $71.03$ & $71.98$ & $71.89$ \\
\cline{2-6}
 & LwF       & $\textbf{66.50}$ & $\textbf{73.04}$ & $73.33$ & $73.20$ \\

\hline
\end{tabular}
\label{tab:merged_results}
\end{table}

\subsection{Experimental Setup}

\textbf{Dataset.} We conduct experiments on CIFAR-100 \cite{cifar}, a widely-used image classification benchmark consisting of 60,000 32×32 RGB images across 100 fine-grained classes. The dataset is split into 50,000 training images and 10,000 test images.\\
\textbf{Architectures.} We adopt the same architectures as in \cite{icml25}. We evaluate two complementary architectures: ResNet-34 \cite{resnet} and MSDNet \cite{msdnet}. For ResNet-34, we augment the standard architecture with 8 internal classifiers positioned at layers $\{2, 4, \cdots, 16\}$, where each exit consists of an SDN-type pooling \cite{sdn}, except for the last classifier, which uses an adaptive average pooling, followed by a linear classifier. For MSDNet, we employ the CIFAR variant with 7 blocks.\\
\textbf{Baselines:} All baseline accuracy scores are taken from \cite{icml25}. We primarily compare our method against three sequential training baselines: disjoint, branch-wise, and separate training, and include joint training for completeness. For our proposed approach, we set $\lambda=100$ and $\rho=0.7$ for ResNet34, and $\lambda=1000$ and $\rho=0.2$ for MSDNet, unless stated otherwise. \\
\textbf{Training Details.} We first pretrain each backbone network with only the final exit using conventional supervised training. Then, following \cite{icml25}, for ResNet-34, we use a batch size of 128, a learning rate of $5 \cdot 10^{-4}$, and no weight decay. For MSDNet, we use a batch size of 512 and a learning rate of $10^{-3}$ with no weight decay. Both models employ CutMix and Mixup augmentation strategies and early stopping with a patience of 50 epochs. After pretraining the backbones, we add ICs sequentially from shallow to deep, applying our proposed regularization at each stage. 

\subsection{Results and Discussion}

Table~\ref{tab:merged_results} presents top-1 accuracy across exits corresponding to different computational budgets ($25\%$, $50\%$, $75\%$, and $100\%$ FLOPs) for our two EENN models on CIFAR-100. Our proposed method consistently outperforms all baseline sequential training approaches. Compared to the strongest sequential baseline (separate training), both EWC and LwF achieve substantial improvements, particularly at low budgets: $+4.25\%$ and $+4.46\%$ at $25\%$ budget, respectively. This demonstrates that our approach effectively balances early exit specialization with deep representation learning. 

The comparison across architectures highlights the interplay between regularization type and backbone topology. In ResNet34, both EWC and LwF improve early-exit performance, with EWC slightly outperforming LwF at the deepest exit, suggesting that parameter-level stabilization is sufficient in sequential residual blocks. The skip connections facilitate feature reuse, so anchoring parameters preserves shallow-exit performance without constraining deeper layers. In contrast, MSDNet's densely connected, multi-scale structure links many layers to multiple exits. Here, LwF consistently outperforms EWC across both early and late exits, indicating that output-distribution regularization better accommodates functional dependencies among exits. By constraining predictions rather than parameters, LwF maintains early-exit accuracy while allowing the backbone to adapt flexibly for later classifiers, effectively balancing stability and plasticity in a highly entangled multi-exit architecture.

Furthermore, we consistently observe improvements over joint training at earlier exits while maintaining comparable performance at later exits. This difference stems from the distinct optimization dynamics of the two approaches. Joint training drives all exits toward a shared compromise solution, whereas our method progressively increases regularization as new exits are added, limiting deeper layers' ability to modify earlier representations. Although this can slightly constrain later adaptation, it yields substantially stronger early exits, aligning with the core goal of EENNs. Further gains may be possible by dynamically scheduling the regularization strength to better balance early and late exits.

\begin{table}[t!]
\centering
\caption{Ablation study on MSDNet for different hyperparameters. \textbf{Reg.} stands for Regularization and defines which type of regularization is used. \textbf{W.U} stands for Warm-up and defines whether the backbone is pretrained.}
\small
\begin{tabular}{c|c|c|c|c|c|c|c}
\hline
Reg. & $\lambda$ & $\rho$ & W.U & $25\%$ & $50\%$ & $75\%$ & $100\%$ \\
\hline
\multirow{4}{*}{EWC}
 & $10$   & -- & True  & $58.33$ & $69.27$ & $73.15$ & $73.39$ \\
 & $10^2$  & -- & True  & $62.54$ & $71.05$ & $72.59$ & $72.43$ \\
 & $10^3$ & -- & True  & $65.10$ & $71.03$ & $71.98$ & $71.89$ \\
 & $10^3$ & -- & False & $66.11$ & $68.91$ & $67.25$ & $66.93$ \\
\cline{1-8}

\multirow{4}{*}{LwF}
 & -- & $0.2$ & True  & $66.50$ & $73.04$ & $73.33$ & $73.20$\\
 & -- & $0.5$ & True  & $68.57$ & $72.63$ & $72.56$ & $72.44$ \\
 & -- & $0.7$ & True  & $67.51$ & $72.74$ & $73.07$ & $72.85$ \\
 & -- & $0.7$ & False & $67.49$ & $70.04$ & $68.99$ & $68.67$ \\
\hline
\end{tabular}
\label{tab:ablation}
\end{table}

To better understand the impact of our design choices, we conduct an ablation study on MSDNet (see \Cref{tab:ablation}). Our ablation explores two main dimensions: (i) the regularization strengths, namely $\lambda$ and $\rho$, and (ii) the impact of an initial pretraining, or warm-up, phase where the full model is trained while optimizing only the final classifier. Across both methods, the strength of the regularization plays a central role in controlling the stability–plasticity trade-off: stronger regularization protects early exits but reduces the backbone's flexibility for deeper layers, while weaker regularization allows adaptation at the cost of shallow-exit performance. The key difference lies in how this constraint is applied. EWC rigidly anchors parameters important for earlier exits, stabilizing low-level features but limiting the backbone's ability to reorganize representations for deeper classifiers. This explains why early-exit performance improves steadily with larger $\lambda$, while high-budget exits are slightly constrained. In contrast, LwF constrains the outputs of earlier exits rather than the parameters themselves, allowing internal representations to adjust as long as the output distribution is preserved. Moderate $\rho$ achieves a balance, maintaining early-exit accuracy while enabling deep exits to benefit from flexible feature adaptation, whereas high $\rho$ overemphasizes shallow classifiers and reduces high-budget performance. Furthermore, backbone warm-up further interacts with both methods by providing an initial alignment of features across layers before regularization is applied. Without a warm-up, both methods struggle: EWC may freeze suboptimal parameters, and LwF may over-constrain features that are not yet meaningfully structured, leading to lower accuracy across all exits. Overall, this ablation study validates the effectiveness of our proposed two-stage strategy: pretraining the final exit establishes rich hierarchical features, and selective regularization allows each exit to adapt effectively without compromising the network's full representational capacity.

\begin{figure}
    \centering
    \includegraphics[width=\linewidth]{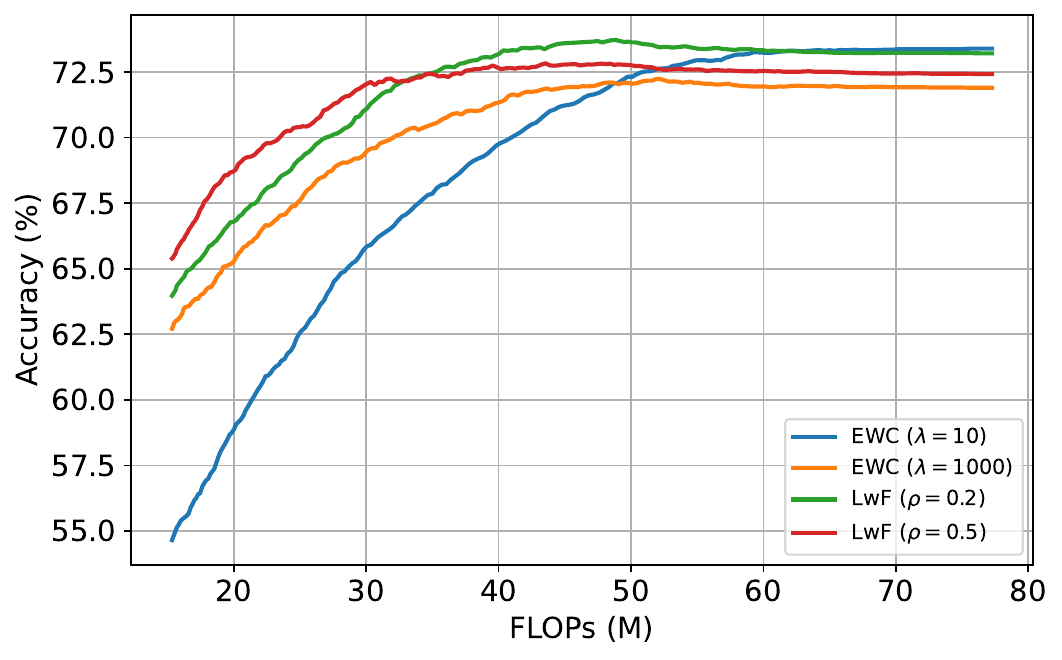}
    \caption{Top-1 accuracy given different FLOPs budgets for MSDNet.}
    \label{fig:acc_flops}
\end{figure}

Figure~\ref{fig:acc_flops} follows the budgeted classification approach (e.g., \cite{eccv24, icml25}) by evaluating the top-1 accuracy of a subset of configurations on MSDNet within a given computational budget. EWC with $\lambda=10$ (blue curve) exhibits delayed accuracy growth, starting at only $55\%$ and requiring approximately 40M FLOPs to reach $70\%$ accuracy, nearly double the computation needed by well-regularized methods. This indicates that weak regularization fails to protect early-layer representations and forces reliance on deeper layers, thus heavier computational load. In contrast, stronger regularization enables substantially faster convergence: LwF variants and well-tuned EWC configurations reach $70\%$ accuracy at roughly 20-25M FLOPs, yielding about a 2x computational speedup.

\begin{figure}
    \centering
    \includegraphics[width=\linewidth]{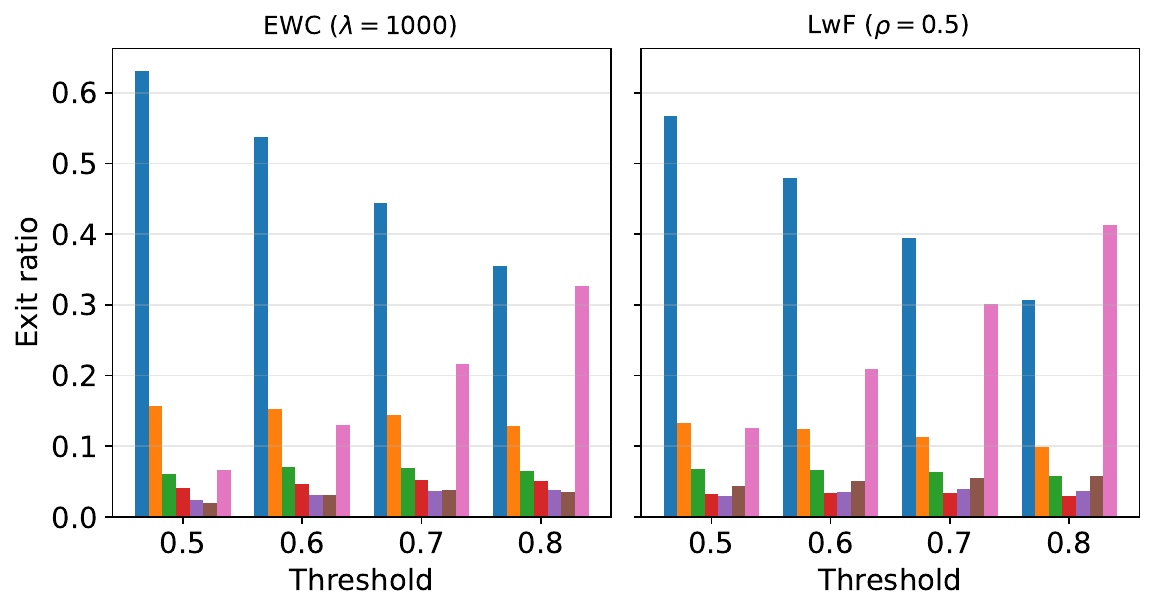}
    \caption{Exit ratios of MSDNet on the CIFAR-100 test set for different thresholds for EWC ($\lambda=1000$) and LwF ($\rho=0.5$) models. For each threshold, exits 1 through 7 are shown from left to right.}
    \label{fig:ex_rat}
\end{figure}

Finally, Figure~\ref{fig:ex_rat} further analyzes how confidence thresholds control the dynamic behavior of MSDNet by reporting exit ratios across all intermediate classifiers. As the threshold increases, predictions are progressively shifted toward deeper exits, indicating that harder samples are routed to later classifiers with richer representations. Comparing EWC ($\lambda=1000$) and LwF ($\rho=0.5$) reveals a systematic difference: LwF exits fewer samples at early classifiers under the same thresholds, while EWC commits earlier. Interpreted together with \Cref{fig:acc_flops}, where LwF achieves higher accuracy at substantially lower FLOPs, this shows that LwF provides a higher effective speedup despite exiting later on average. This behavior suggests that LwF learns more accurate but less confident early classifiers, deferring decisions more often, whereas EWC produces earlier but less accurate exits.

\section{Conclusion}
In this work, we propose a novel strategy for training EENNs to mitigate the degradation of previously learned representations caused by incremental exit addition. By incorporating parameter-level protection via EWC and output-distribution consistency via LwF, we establish a principled sequential training framework that balances exit specialization with knowledge preservation. Our experimental study demonstrates that the proposed approach consistently outperforms existing sequential training methods and achieves better early exit accuracy than joint training. Future work will investigate adaptive regularization scheduling and more structured protection mechanisms, e.g., at the level of convolutional kernels or attention heads.

\newpage

{\fontsize{8.6}{10.3}\selectfont
\bibliographystyle{IEEEbib}
\bibliography{refs,strings}
}

\end{document}